# Self-supervised learning of imaging and clinical signatures using a multimodal joint-embedding predictive architecture

Thomas Z. Li[1,2], Aravind R. Krishnan[3], Lianrui Zuo[3], John M. Still[4], Kim L. Sandler[5], Fabien Maldonado[6], Thomas A. Lasko[4], Bennett A. Landman[1,3,4,5]

[1]Department of Biomedical Engineering, Vanderbilt University, Nashville, TN

[2]Medical Scientist Training Program, Vanderbilt University, Nashville, TN

[3]Department of Electrical and Computer Engineering, Vanderbilt University, Nashville, TN

[4]Department of Biomedical Informatics, Vanderbilt University, Nashville, TN

[5]Department of Radiology and Radiological Sciences, Vanderbilt University Medical Center, Nashville, TN

[6]Department of Medicine, Vanderbilt University Medical Center, Nashville, TN

## Abstract

The development of multimodal models for pulmonary nodule diagnosis is limited by the scarcity of labeled data and the tendency for these models to overfit on the training distribution. In this work, we leverage self-supervised learning from longitudinal and multimodal archives to address these challenges. We curate an unlabeled set of patients with CT scans and linked electronic health records from our home institution to power joint embedding predictive architecture (JEPA) pretraining. After supervised finetuning, we show that our approach outperforms an unregularized multimodal model and imaging-only model in an internal cohort (ours: 0.91, multimodal: 0.88, imaging-only: 0.73 AUC), but underperforms in an external cohort (ours: 0.72, imaging-only: 0.75 AUC). We develop a synthetic environment that characterizes the context in which JEPA may underperform. This work innovates an approach that leverages unlabeled multimodal medical archives to improve predictive models and demonstrates its advantages and limitations in pulmonary nodule diagnosis.

## 1. Introduction

Although certain predictive models have demonstrated high accuracy in predicting pulmonary nodule diagnosis, this performance does not generalize well across different institutions and clinical settings[1]. These models tend to perform poorly on examples that deviate from their training distribution, highlighting an urgent need to address sources of heterogeneity that drive such performance variability. One promising approach is to model this heterogeneity directly using self-supervised learning on large-scale medical archives. We hypothesize that self-supervised representations learned from large unlabeled datasets can capture transferable features that help regularize lung cancer classification in smaller, labeled cohorts. In this study, we focus specifically on addressing intra-site heterogeneity at our home institution.

Previous work in self-supervised learning from multimodal medical data primarily span three strategies: contrastive, self-prediction, and generative, the last of which sees use in disease diagnosis (or other classification tasks)[2,3]. Contrastive Language-Image Pretraining (CLIP) attempts to align pairs of images and corresponding expert-annotations within a unified representation space by optimizing for some measure of similarity between true pairs while contrasting arbitrary pairs. CLIP excels in radiology and pathology where image-caption pairs are available[4–7]. When imaging and non-imaging data are not semantically matched, which is the case of most clinical data outside of radiology and pathology reports, the self-prediction approach involves masking parts of the input and teaching the model to reconstruct the full signal from the unmasked context. Depending on the masking strategy, self-prediction can teach models to predict some missing distribution from a sequence[8] or missing patches of an image[9,10]. As recent addition to self-prediction approaches, joint-embedding predictive architecture (JEPA)[11] challenges the model to predict randomly masked portions of the input given the surrounding context. The key difference with this approach is that the model is not asked to reconstruct the input space, but rather the reconstruction loss is measured in the latent space of the encoder (Figure 1).

Regardless of the approach, self-supervised learning with 3D medical images is extremely challenging when conducted along-side non-imaging data. As a result, almost all approaches, including most foundation models released to this date, are limited 2D imaging and some semantically related text [12]. To advance predictive model generalization, evaluation of self-supervised approaches must include longitudinal multimodal data that realistically reflect the clinical routine. To this end, we innovate an extension of JEPA for self-supervision and evaluates its utility in lung cancer diagnosis for an external cohort. To power pretraining, we create a new cohort (n=5518) of chest CTs and EHR-derived clinical signatures[13] which captures patients from imaging archives who have at least one chest CT and a pulmonary nodule coded for at any point in their record from Vanderbilt University Medical Center (VUMC). We compare JEPA with a fully supervised counterpart, as well as single modality models utilizing longitudinal imaging or longitudinal signatures. Surprisingly, we find that multimodal JEPA outperforms other approaches on an internal test set, but performs on par with longitudinal imaging on the external cohort. To understand where our approach falls short, we create a synthetic dataset and repeat the comparative evaluations. Within this context, we find that JEPA does not significantly enhance finetuned classification compared to a purely supervised approach.

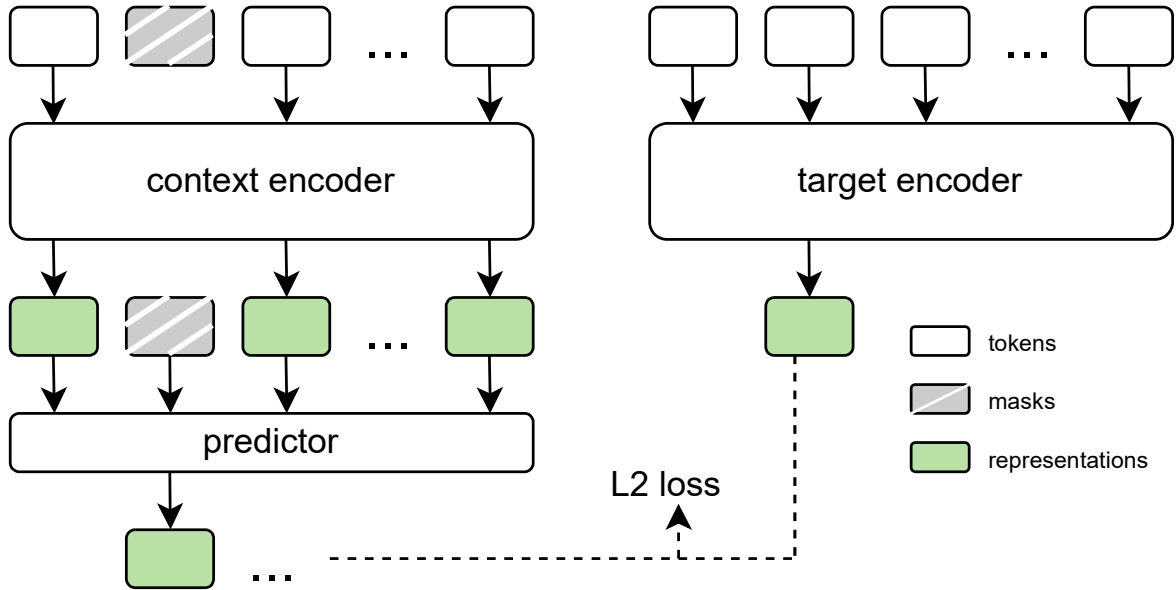

**Figure 1.** Training scheme for joint-embedding predictive architectures. For each subject 15% of tokens are randomly and non-contiguously masked. The remaining 85% of tokens are fed into a context encoder which outputs a context representation of the input and maintaining the sequence structure. Next, a lightweight predictor network takes the full sequence of context tokens along with masked tokens and predicts the representation of the masked tokens. The target representations are the outputs of a target encoder that is given the unmasked sequence. The weights of the target-encoder are updated after each iteration to be the exponential moving average of the context encoder. We employ a mean-squared error loss to measure the distance between the predicted and target representations.

## 2. Methods

### 2.1 Data

A large effort of this study was the creation of NoduleVU, new multimodal dataset from VUMC that included imaging with corresponding EHRs acquired under Vanderbilt IRB #140274. Since the intention of this dataset was to capture intra-site heterogeneity, we prioritized size and diversity. We broadly selected VUMC patients who had a billing code for a pulmonary nodule at any point and at least one CT session from ImageVU. Due to the growing infrastructure and data-sharing protocols, this imaging archive is an incomplete and non-contiguous sample of all imaging acquired at VUMC. For the purposes of this study, we will assume images are missing at random conditioned on parameters in the EHR we are able to observe[14].

### 2.2 Image Preprocessing

All images acquired underwent an extensive quality assurance pipeline.

1. Beginning with 55,888 CT sessions, we removed series with a slice thickness greater than 5mm, excluded series with missing slices, and converted DICOM to NIfTI.

2. Observing that a large minority of series were not of the chest, we ran a body part regression algorithm [15] to categorize the field of view as spanning the head, chest, abdomen, none, or a combination of the three. CTs without the chest categorization were excluded (https://github.com/MIC-DKFZ/BodyPartRegression).

3. We performed a visual assessment of all scans that passed the preceding steps, removing unusable scans. Reasons for discard included non-standard body orientation, abnormal anatomy (such as lobectomy from surgery), and visible artifact occluding the lung field.

4. For each session, we selected the scan best approximating the Siemens "soft" B30f kernel (Figure 2). Selection was automated by selecting the medium slice and computing the energy along the max radius $r$ of its discrete Fourier transform. The "softest" kernel was assumed to be the scan with the smallest integral over the circumference:

$$\int_0^{2\pi} \nu(\theta)^2 r d\theta$$

where $\nu(\theta)$ denotes frequency at angle $\theta$.

5. All scans were axially cropped to the lung fields using TotalSegmentator[16] with a randomized 1-5cm buffer.

27,170 CT scans across 5518 subjects were identified as usable after quality assurance.

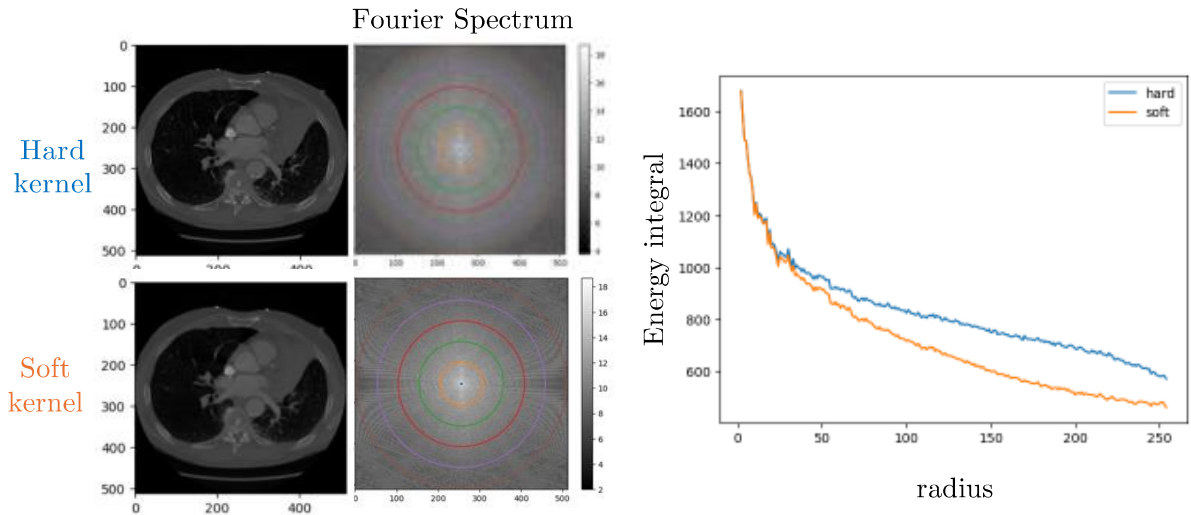

**Figure 2.** We preferred scans with soft kernels over hard kernels when selecting a single scan from multiple within a session. Integrating the energy of the Fourier spectrum over a 2D circumference (colored) provides a simple measure of how "soft" or smooth a reconstruction kernel is.

## 2.3 EHR Preprocessing

Integrating tabular EHR and imaging data longitudinally is challenging. We use the Li et al. approach[17] to synchronize and encode the longitudinal multimodal data in this study.

First we used the Lasko et al. pipeline[13,18] to impute EHR variables synchronized with scan dates. Specifically we built longitudinal curves from variables spanning diagnostic codes, procedures, medications, and labs. These curves require at least two instances of a variable to impute values within 100 days after the latest instance. The variable is considered missing outside of this range. For any given cross-section, values are only imputed from variable instances occurring at or before the cross-section. We sampled the closest available cross-section of each longitudinal curve at or before each scan acquisition date.

For each cross-section, the full set of EHR variables as a vector was then projected on to a $\mathbb{R}^{2000}$ feature space we call clinical signatures. This feature space was inferred using Independent Component Analysis in a previous study[19] of 9195 variables from 269,099 longitudinal records from patients with general pulmonary conditions. EHR variables projected in this way are called expressions of clinical signature. If the patient had no usable data to build curves from before a scan date, we considered the expression missing. In this way we computed 24,303 EHR-derived clinical signature expressions from 5518 patient. Each patient was represented as a set of cross-sections, and each cross-section was a (scan, expression) pair or a (scan,) if the expression was missing.

## 2.4 Clinical Cohorts

In addition to NoduleVU, we reproduce the LiVU-SPN and MCL-VUMC cohorts from a previous paper[1]. Briefly, LiVU-SPN consists of VUMC patients with a pulmonary nodule code and no prior history of any cancer. MCL-VUMC consist of prospectively enrolled VUMC patients with incidentally-detected pulmonary nodules between 6-30mm. Patients in both cohorts were labeled as positive if they were diagnosed with lung cancer after 2 years and negative otherwise. The label was assigned based on diagnostic codes for LiVU-SPN and follow up biopsy reports for MCL-VUMC. A notable distinction between the two cohorts is that LiVU-SPN is a subset of NoduleVU, while MCL-VUMC is a true external test set completely disjoint from both cohorts (Table 1).

**Table 1.** Cohort Characteristics

| Dataset | NoduleVU | LiVU-SPN | MCL-VUMC |
|---|---|---|---|
| Patients | 5518 | 218 | 265 |
| Scans | 27170 | 514 | 621 |
| Expressions | 24303 | 512 | 594 |
| % cancer | N/A | 12% | 69% |
| Stage of training used | JEPA pretraining | Fine-tune | Fine-tune |

## 2.5. Experiments with Clinical Data

In NoduleVU, each patient is represented by a sequence of CT scans and clinical signature expressions sampled at the scan date. A feature vector was computed for each CT scan using Sybil [20], an open-sourced ResNet trained on multi-institute lung screening cohort. These image-derived feature vectors and expressions were linearly projected into a joint space in $\mathbb{R}^{512}$ with learnable matrices, each as a separate token. Time encodings corresponding to the cross-section timestamp and modality embeddings of the same dimension were added to each token. Added together these formed in the input sequence to the JEPA model. We trained on NoduleVU with a 15% mask ratio and during training the context tokens were represented with the time encodings added to the modality embeddings. We set the learning rate to 0.001 with a cosine annealing schedule, batch size to 1000 and employed Adam optimization.

We evaluated multimodal JEPA through its fine-tune performance on an internal test set (LiVU-SPN) and external test set (MCL-VUMC). The fine-tune model was the context encoder initialized with JEPA-trained weights preceding an MLP classifier. We compared this approach to a model with identical architecture but trained in a purely supervised manner from random weights. We also evaluated single modality baselines with a longitudinal imaging model using the same Sybil feature vectors as well as a longitudinal expression model. Both single modality baselines were time-distance transformers with the same architecture as the JEPA context encoder.

## 3. Clinical Results

The multimodal models (JEPA: 0.905 AUC, supervised: 0.879 AUC) outperformed single modality models on LiVU-SPN (Imaging: 0.733, Expressions: 0.615). In this test set, multimodal JEPA was the best performer. Surprisingly, JEPA struggled to differentiate itself on VUMC-MCL at 0.724 AUC compared to the longitudinal imaging model at 0.753 AUC. The multimodal supervised model also performed poorly on the external test set at 0.696 AUC (Table 2).

**Table 2.** AUC of 2-year lung cancer classification on internal and external test sets

| Dataset | LiVU-SPN* | MCL-VUMC[#] |
|---|---|---|
| Longitudinal Expressions | 0.615 | 0.624 |
| Longitudinal Imaging | 0.733 | **0.753** |
| Multimodal Supervised | 0.879 | 0.696 |
| Multimodal JEPA | **0.905** | 0.724 |

*internal test, [#]external test

## 4. Synthetic Experiments

After observing mixed results in the clinical experiments, we designed a synthetic cohort to understand in what context JEPA underperforms. In this cohort, we devised four independent causal variables from which data and labels were generated.

**Table 3.** Synthetic Datasets

| Dataset | P(G1), P(D1) | P(G2), P(D2) | Experimental stage | Size | Modalities |
|---------|--------------|--------------|--------------------|------|-----------|
| $U$ | Bern(0.5) | Bern(0.5) | Unsupervised pretraining | 40,000 | Synthetic nodules, Synthetic expressions |
| $S$ | Bern(0.5) | 0 | Supervised pretraining | 40,000 | Cross-sectional synthetic nodules, Synthetic expressions |
| $F$ | Bern(0.5) | Bern(0.5) | Finetune | 50 to 8000 | Synthetic nodules, Synthetic expressions |
| $T$ | Bern(0.5) | Bern(0.5) | Test | 2000 | Synthetic nodules, Synthetic expressions |

## 4.1. Synthetic Data

Under the assumptions of a dataset composed of imaging and non-imaging modalities, we devised a data generating process that output 2D images with superimposed nodule-like lesions and $m$-dimensional expressions. Each subject in our synthetic study was modeled using four independent $Bernoulli\left(\frac{1}{2}\right)$ random variables $G1, G2, D1, D2$ that causally influenced the subjects' images and expressions through a pseudo-random data-generating process. Formally, we sampled $n$ i.i.d. samples of causal variables $G1, G2, D1, D2$ and generated a corresponding set of images $I$ and expressions $E$. We also describe how these data are represented when variables $G2$ and $D2$ are missing.

**Synthetic Images.** The following steps were used to generate synthetic imaging data, $I$, from the causal variables $G1$ and $G2$.

1. Sample two nodule features for each subject: initial size, $s$, conditioned on $G1$ and growth rate, g, conditioned on $G2$.

$$s \sim \begin{cases} N(1, 0.06), & if\ G1 = 0 \\ N(3, 0.06), & if\ G1 = 1 \end{cases} \qquad g \sim \begin{cases} N(0.5, 0.06), & if\ G2 = 0 \\ N(1.5, 0.06), & if\ G2 = 1 \end{cases}$$

2. For each subject, draw a 5-image series of a nodule with initial size $s$ and a growth rate of $g$, randomly sampling the points in time from which to draw each cross-section, with the first image starting at time zero. Each subsequent image has a corresponding timestamp denoting the time passed from the first image (Figure 4).

3. We follow the Tumor-CIFAR pipeline (https://github.com/MASILab/tumor-cifar) which superimposes nodule-like lesions of randomly sampled morphologies on top of images randomly sampled from the CIFAR-10 dataset [21]. A random translation and rotation was applied to each nodule series. In addition, salt and pepper noise is added.

4. If $G2$ is for subject $i$, $I_i$ is equal to the single image at time zero.

**Synthetic expressions.** We generated synthetic expressions as a sparse and high-dimensional projection of causal variables $D1$ and $D2$. Let $D$ be a $2 \times n$ matrix representing the $n$ choices for $D1$ in the first row and the $n$ choices for $D2$ in the second row. Subtracting by 0.5 centered the values in $D$ around 0, which was the fill-in value for a choice of $D2$ that was missing.

Next, let $A$ be a sparse binary $m \times 2$ matrix where each element is sampled from a $Bernoulli(0.01)$ distribution. The product $AD$ samples $D1$ and $D2$ at 0.01 probability, producing a sparse feature vector $\in R^m$ for each subject.

Finally, we add gaussian noise weighted with parameter $\beta \in (0,1)$. In summary, the synthetic expression data, $E$, is generated as a set of feature vectors $\in R^m$ represented by a $n \times m$ matrix from the linear mixture of $A$ and $D$ with added noise $\epsilon \sim N(0,0.5)$.

$$E = \beta(A(D - 0.5)^{\mathsf{T}}) + (1 - \beta)\epsilon$$

After experimentation, the choice of $\beta = 0.01$ was made to be minimal while ensuring the model's performance was not substantially degraded. We set m=128 to half of our model's attention head dimension.

**Synthetic Labels.** In addition to $I$ and $E$, a set of binary labels $L$ was deterministically generated from causal variables with the following rule:

$$L = \begin{cases} 1, & if \ (G1, G2, D1, D2) \in \{(1,0,1,0), (0,1,0,1)\} \\ 0, & else \end{cases}$$

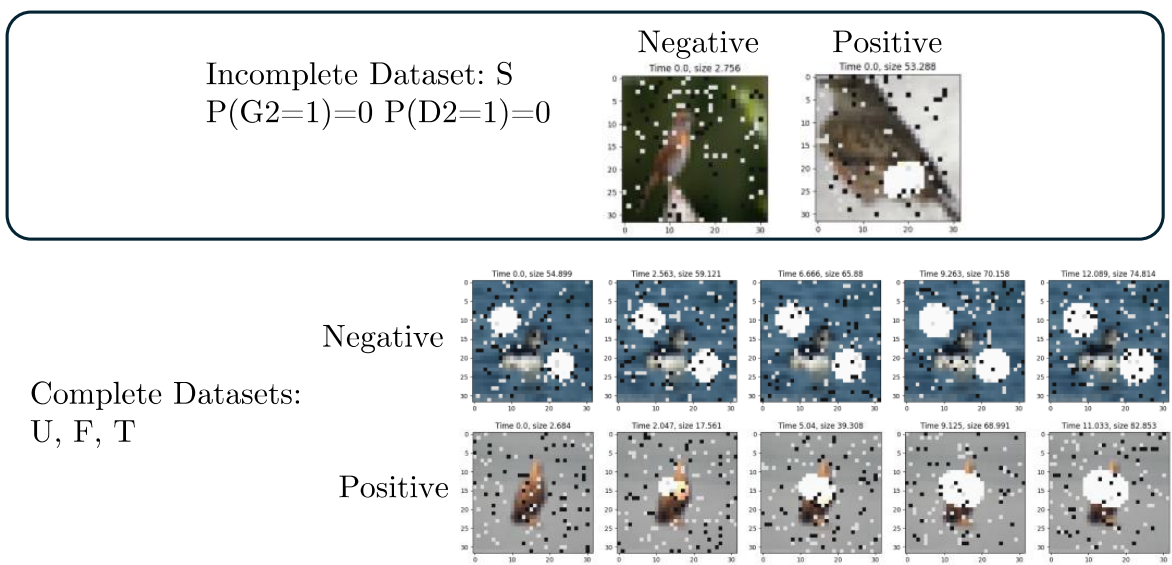

**Figure 4.** A series of synthetic nodules were generated from latent variables G1 and G2. Since the incomplete dataset, S, was missing G2 which determined growth rate, we only provided the first cross-section. Randomly salt and pepper noise and random translation was applied to all images.

## 4.2. Synthetic Datasets

We modeled a scenario where unsupervised dataset contained additional heterogeneity of the causal variables compared to a labeled dataset with limited casual diversity. Thus we generate three "complete" datasets: an unlabeled dataset $U = \{I^U, E^U\}$, a labeled finetuning dataset $F = \{(I^F, E^F, L^F)\}$, and a test dataset $T = \{(I^T, E^T, L^T)\}$. We also generate an "incomplete" labeled dataset $S = \{(I^S, E^S, L^S)\}$. $I^U, I^S, I^F$, and $I^T$ form a partitioning of $I$ (Table 3). The incomplete dataset differs from the complete datasets both in its data availability and its underlying distribution. In $S$, the causal variables $G2$ and $D2$ are set to a constant $G2 = 0, D2 = 0$. In addition, these variables are unobserved by the data generating process.

Following the data generating process described in the previous section, each instance in $I^U$ is a 5-image series of a nodule with an initial size and growth rate conditioned on $G1$ and $G2$, whereas each instance in $I^S$ is a single image at time zero of a nodule with initial size conditioned on $G1$ only. Similarly, $E^U$ is a rank 2 matrix conditioned on $D1$ and $D2$, whereas $E^S$ is a rank 1 matrix conditioned on $D1$ only. $L^S$ was generated following the predefined rule from random $G1, D1$ and constant $G2 = 0, D2 = 0$.

### 4.2. Expected Performance

The question we pose in this synthetic study is as follows: does unsupervised training with JEPA on a complete dataset lead to improved performance after finetuning compared to supervised pre-training on an incomplete dataset? Specifically, we hypothesize that a model trained on $U, F$ to outperform one trained on $S, F$ when both are evaluated on a separate test set $T$.

This hypothesis is predicated on the assumption that we have correctly initialized S as an incomplete dataset. Here we calculate the expected AUC of a predictive model trained on $S$ when evaluated on a test set $T = \{I^T, E^T\}$. This is a scenario where a model is trained on an incomplete dataset, but is evaluated on a complete dataset. We assume that a model trained on $S$ would learn a scoring function that perfectly separates the instances of $G_1 = 1$ and $D_1 = 1$ from all other distributions:

$$f(G_1, D_1) = 1 \text{ if } G_1 = 1 \text{ and } D_1 = 1, \text{ else } 0$$

In practice, the predictions of a neural network would be real-valued between [0,1], but we assume a binary scoring function to establish an upper bound under the assumptions of a model that learns to predict the labels in $S$ with perfect confidence.

Next, we evaluate $f$ on $T$ with respect to all possible combinations of $(G1, G2, D1, D2)$ and classify each prediction as a true/false positive/negative.

$$(G1, G2, D1, D2) = (1,0,1,0) \ \rightarrow \ f(G_1, D_1) = 1, \text{ true positive}$$

$$(G1, G2, D1, D2) = (0,1,0,1) \ \rightarrow \ f(G_1, D_1) = 0, false\ negative$$

$$(G1, G2, D1, D2) \in \{(1,1,1,1), (1,0,1,1), (1,1,1,0)\} \ \rightarrow \ f(G_1, D_1) = 1, \text{ false positive}$$

$$(G1, G2, D1, D2) \in \{11\ combinations\ not\ included\ above\} \ \rightarrow \ f(G_1, D_1) = 0, true\ negative$$

For discrete scores, AUC is defined as the probability that the score from a randomly drawn positive example exceeds a score from a randomly drawn negative example, where ties are broken uniformly at random[22]. Let $X^+$ be a random positive example and $X^-$ be a random negative example.

$$AUC = P\big(f(X^+) > f(X^-)\big) + \frac{1}{2}P\big(f(X^+) = f(X^-)\big)$$

Out of 14 possible negative examples, 11 true negatives score below a 1 if the positive example is a true positive. In other words,

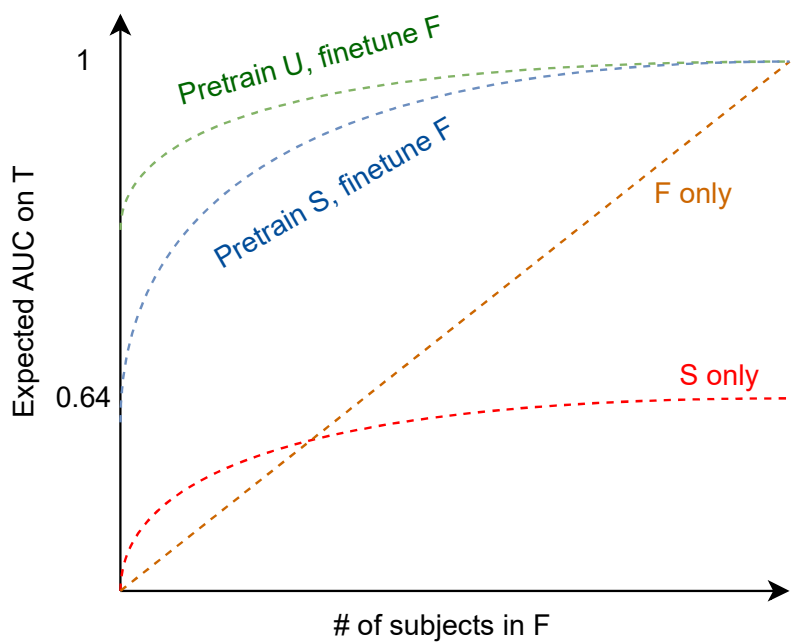

**Figure 3.** Expected AUC for test set T for different training scenarios. Assuming the latent variables are perfectly learnable from the generated data, models trained on complete dataset F should eventually converge to perfect performance when given enough labeled examples. On the other hand, models trained on incomplete dataset S have an upper bound in performance. We hypothesize that a model pretrained on U via self-supervision will perform better with less labeled data compared to a model trained on incomplete dataset S.

$$P\big(f(X^+) > f(X^-)\big) = P(f = 1|X^+)P(f = 0|X^-) = \frac{1}{2} \cdot \frac{11}{14}$$

In the case of $f(X^+) = f(X^-)$, we consider two cases: when the randomly drawn positive example is a true positive, resulting in a score of 1, and when the drawn positive example is a false negative, resulting in a score of 0. In the first case, 3 false positives out of 14 negative examples match the positive example's score of 1. In the second case, 11 out of 14 examples match the positive example's score of 0. Therefore:

$$P\big(f(X^+) = f(X^-)\big) = P(f = 1|X^+)P(f = 1|X^-) + P(f = 0|X^+)P(f = 0|X^-) = \frac{1}{2} \cdot \frac{3}{14} + \frac{1}{2} \cdot \frac{11}{14}$$

Finally we have

$$AUC = \frac{1}{2} \cdot \frac{11}{14} + \frac{1}{2} \left( \frac{1}{2} \cdot \frac{3}{14} + \frac{1}{2} \cdot \frac{11}{14} \right) = 0.642$$

In theory, a model trained on $S$ would have an expected upper bound of 0.642 AUC on test set $T$.

Having established an expected classification performance for a model trained on S alone, we ask whether unsupervised training on a complete dataset improves on supervised pretraining on an incomplete dataset. Naturally the models' test performance would depend on the size of the datasets. Fixing the size of $S, U, T$, we hypothesize that the expected AUC will correlate positively with the size of $F$. Given extreme case of large $F$, both models trained on $S$ and $U$ would have the potential to achieve perfect performance on $T$. However, with limited sized $F$, we expect a model trained on $U, F$ to outperform one trained on $S, F$ due to the former having seen the complete distribution. In other words, we hypothesize that JEPA training on $U$ will enable faster performance convergence with respect to the size of $F$ (Figure 3). In the following sections, we study this hypothesis experimentally.

## 5. Synthetic Results

We pretrained a JEPA encoder using dataset $U$ (n=40,000). We finetuned this encoder with dataset $F$ of different sizes (n=50-8000) to determine how model performance varied with the number of labels seen. We also trained a supervised model in two stages – first on $S$ (n=40,000) and then on $F$ (n=8000). As described in the previous section, variables $G2$ and $D2$ are missing in $S$, but we also conducted experiments where all variables were missing except $G1$, meaning the input to the model was a single image and no expressions were available. We evaluated the model on test set $T$ (n=2000) as well as a separate test set with the label distribution of $S$ (n=1000).

The architecture of all models resembled those used in clinical experiments but with the size of the model scaled down at all levels. We also trained the ResNet module from random weights without using Sybil. Hyperparameters remain largely unchanged from clinical experiments, except for a substantial increase in batch size scaling with the larger number of examples.

**Table 4.** AUC classification on S and T with different training sets. We set the size of F to 8000 in this round of evaluation.

| Training Strategy | Training Set (Latent variables available) | Test split of S (n=1000) | T (n=2000) |
|---|---|---|---|
| Supervised | *S (G1)* | 0.83 | 0.43 |
| Supervised | *S (G1, D1)* | 0.99 | 0.49 |
| Supervised-finetune | *S* and *F (G1, G2, D1, D2)* | 1.0 | 0.99 |
| JEPA-finetune | *U* and *F (G1, G2, D1, D2)* | N/A | 0.99 |

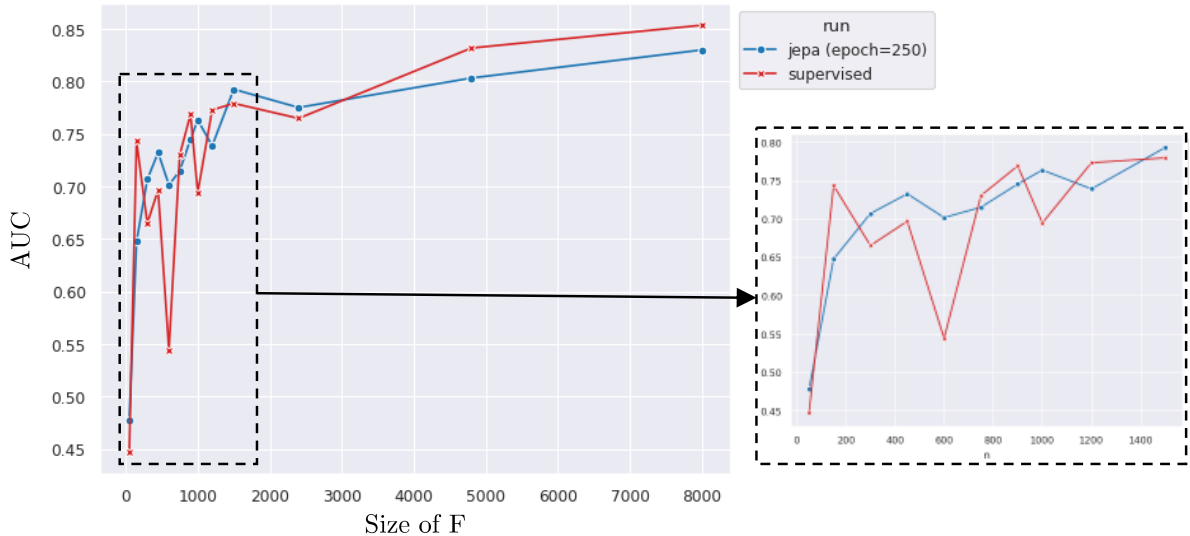

**Figure 5.** Comparing supervised-finetune and JEPA-finetune strategies for at different finetune sizes. We expected JEPA-finetune to outperform the supervised strategy at small dataset sizes, but no significant difference in performance across size of *F* was observed (p=0.635).

Initial evaluation revealed results that aligned with our theoretical projections. Namely, supervised pretraining on an incomplete dataset *S* without finetuning resulted in poor performance on a complete dataset *T*. Finetuning with a complete dataset of *F* bridged this gap as both training with *S, F* and *U, F* resulted in near perfect performance on *T*. These results establish the lower and upper bounds of test performance at no finetuning and finetuning with 8000 examples respectively.

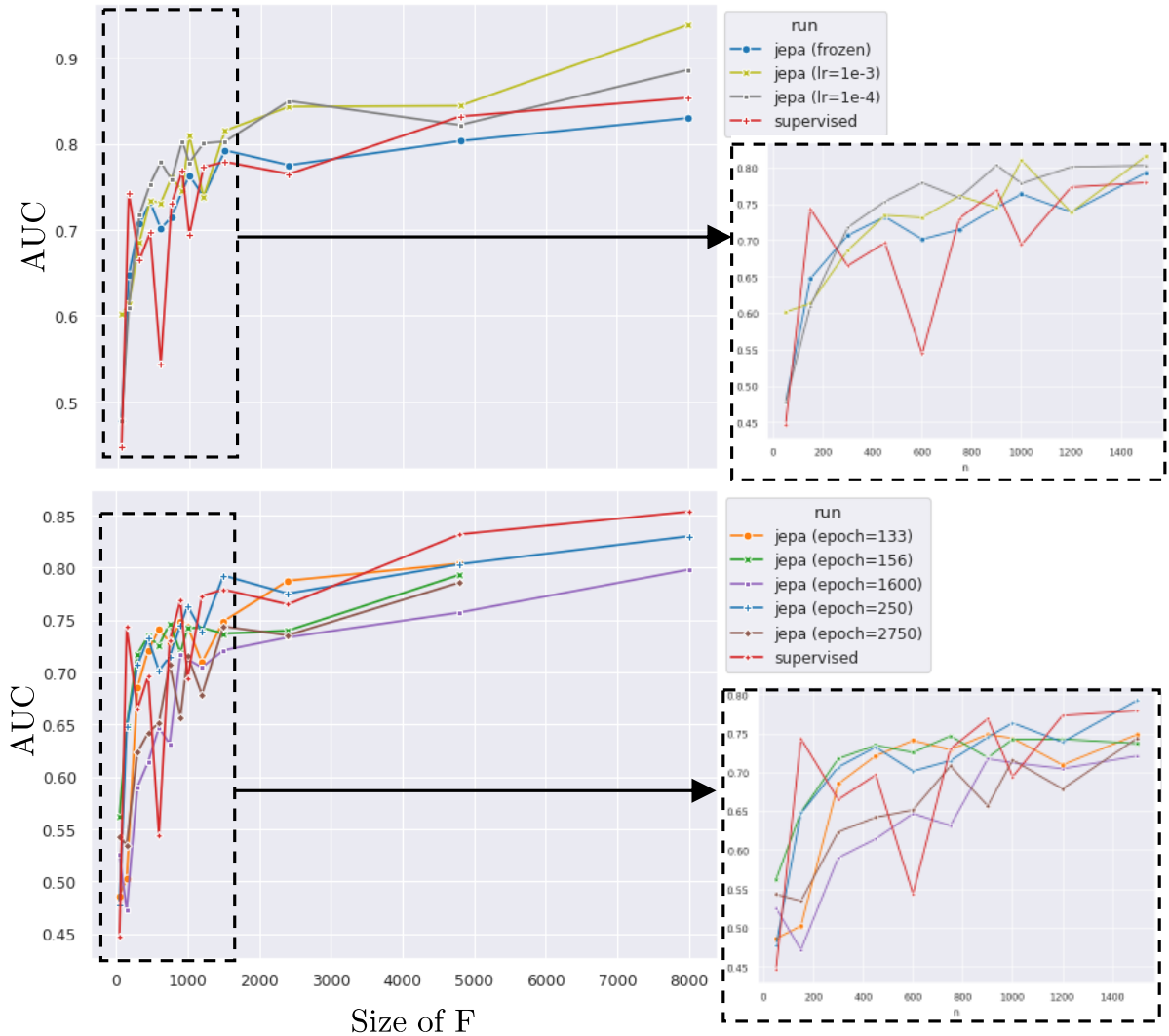

**Figure 6.** Varying the learning rate of the JEPA encoder (top) and JEPA checkpoint choice (bottom) did not significantly improve classification performance, even at small data regimes.

Next, we examined how JEPA-finetune compared with Supervised-finetune at varying sizes of *F*. We repeated the finetuning stage for both training strategies at different dataset sizes ranging from 50 to 8000. Figure 5 directly compares the two training strategies. Treating two model evaluations on the same size of F as a pair, we tested if there was a significant difference in performance across size of *F* using the Wilcoxon signed-rank test. We found no significant difference between the Supervised and JEPA approach with a p=0.635.

Figure 6 (top) analyzed the effect of different learning rates, and Figure 6 (bottom) evaluated the same experiment with different checkpoints in JEPA pretraining. Repeating statistical test between the Supervised approach and each JEPA variant, we found no significant difference in performance across size of *F*.

## 6. Discussion

This study innovates a multimodal extension of JEPA to account for inter-site heterogeneity towards improving the diagnosis of incidental nodules. We curated a new longitudinal and multimodal cohort of over more than 27000 scans and 24000 clinical signature expressions across 5000 subjects that will fuel the study of patients with pulmonary nodules for current and future projects. Evaluation into our approach showed improvement over supervised and single modality baselines in one clinical cohort, but underperformed on a true external cohort compared to a much simpler longitudinal imaging model. Lastly, we devised synthetic experiments to discovery where JEPA pretraining might not be useful.

The underwhelming performance of both the multimodal models on MCL-VUMC suggest that integrating imaging and clinical signatures in this cohort leads to overfitting when a simpler imaging model does not. This is a well-known challenge of multimodal classification [23] and in this case the increased number of parameters relative to the size of the training set may have exacerbated overfitting. During JEPA training, we experienced several instances where a specific choice of hyperparameters led to model collapse, a failure mode where the model learns trivial representations that compute low levels of loss but have few meaningful features (Figure 7). Due to the "black-box" nature of these models, we will not speculate on the exact cause of model collapse but suggest that it likely arises from a fundamental flaw in the approach itself. Even when no obvious model collapse occurred, the training loss converged much faster than expected. The fine-tune results certainly corroborate that, at least in MCL-VUMC, the joint representation is less useful than a simpler single-modality model.

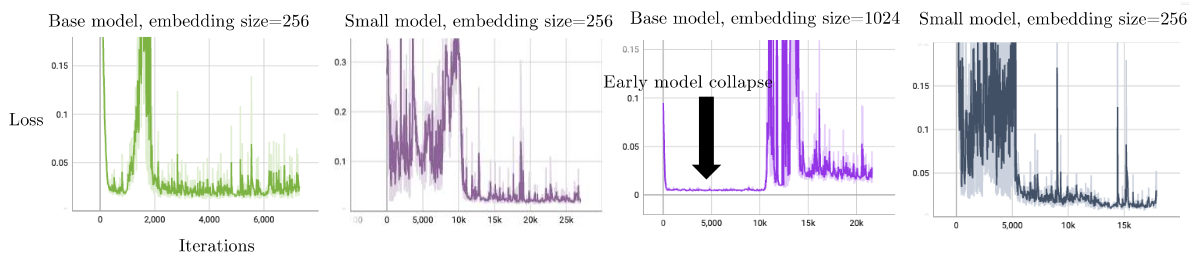

**Figure 7.** JEPA loss over training steps with different hyperparameters converges to zero after 5000-10,000 iterations, equivalent 50-100 epochs. Given complexity of data being modeled, we expect self-supervised learning require longer training times and we suspect some degree of model collapse is occurring. The third example is an especially obvious of example of modal collapse.

We also identified a failure case for the JEPA approach in a synthetic environment where data is generated from independent causal variables. In this setting, we expected JEPA training on dataset $U$ to act as a form of unsupervised dimensionality reduction, compressing the multimodal input into a small number of features correlated with the underlying causal variables—i.e., to learn a representation informative about the data generating process. Given that JPEA was not successful in improving performance over any finetune size, we again suspect some fundamental issue with JEPA leading to model collapse. Although the synthetic data was not intended to resemble real clinical distributions and involved several arbitrary choices in the generative process, the underlying structure was deliberately designed to be learnable. As such, the failure of JEPA to improve performance at any finetuning scale suggests a more fundamental limitation of the method, potentially related to model collapse.

## 7. Acknowledgments

This research was funded by the NIH through F30CA275020, 2U01CA152662, and R01CA253923-02, as well as NSF CAREER 1452485 and NSF 2040462. This study was also funded by the Vanderbilt Institute for Surgery and Engineering through T32EB021937-07, the Vanderbilt Institute for Clinical and Translational Research through UL1TR002243-06, and the Pierre Massion Directorship in Pulmonary Medicine. We utilize generative AI to generate code segments based on task descriptions, as well as to assist with debugging, editing, and autocompleting code. Additionally, generative AI has been employed to refine sentence structure and ensure grammatical accuracy. However, all conceptualization, ideation, and prompts provided to the AI stem entirely from the authors' creative and intellectual efforts. We take full responsibility for reviewing and verifying all AI-generated content in this work.